\documentclass[letterpaper, 10pt, conference]{ieeeconf}  

\IEEEoverridecommandlockouts                              

\usepackage{graphicx} 
\usepackage{wrapfig}
\usepackage{amsmath}
\usepackage{amsfonts}
\usepackage{amssymb}
\usepackage{bbm}
\usepackage{xcolor}

\usepackage{url}
\usepackage{cite}

\usepackage{dsfont}
\usepackage{braket}
\usepackage{booktabs}
\usepackage{color}
\usepackage{colortbl}
\usepackage{mathtools}
\usepackage{multirow}
\usepackage{footmisc}

\usepackage{etoolbox}
\makeatletter
\patchcmd{\@makecaption}
  {\scshape}
  {}
  {}
  {}
\makeatother

\title{\LARGE \bf
Multi-View Dreaming: \\
Multi-View World Model with Contrastive Learning}
\author{Akira Kinose$^{1*}$, Masashi Okada$^{2}$, Ryo Okumura$^{2}$, Tadahiro Taniguchi$^{2,3}$
\thanks{$^{1}$ Akira Kinose is with Innovation Center, Connected Solutions Company, Panasonic Corporation, Japan.}%
\thanks{$^{2}$ Masashi Okada and Tadahiro Taniguchi are with Digital \& AI
Technology Center, Technology Division, Panasonic Corporation, Japan.}%
\thanks{$^{3}$ Tadahiro Taniguchi is also with Ritsumeikan University, College of
Information Science and Engineering, Japan.}%
\thanks{$^{*}$ kinose.akira@jp.panasonic.com}}%

\begin{document}
\maketitle
\thispagestyle{empty}
\pagestyle{empty}

\begin{abstract}

In this paper, we propose Multi-View Dreaming, a novel reinforcement learning agent for integrated recognition and control from multi-view observations by extending Dreaming.
Most current reinforcement learning method assumes a single-view observation space, and this imposes limitations on the observed data, such as lack of spatial information and occlusions. 
This makes obtaining ideal observational information from the environment difficult and is a bottleneck for real-world robotics applications. 
In this paper, we use contrastive learning to train a shared latent space between different viewpoints, and show how the Products of Experts approach can be used to integrate and control the probability distributions of latent states for multiple viewpoints.
We also propose Multi-View DreamingV2, a variant of Multi-View Dreaming that uses a categorical distribution to model the latent state instead of the Gaussian distribution.
Experiments show that the proposed method outperforms simple extensions of existing methods in a realistic robot control task.

\end{abstract}

\section{Introduction}
\begin{figure}[t!]
  \centering
  \includegraphics[width=8cm]{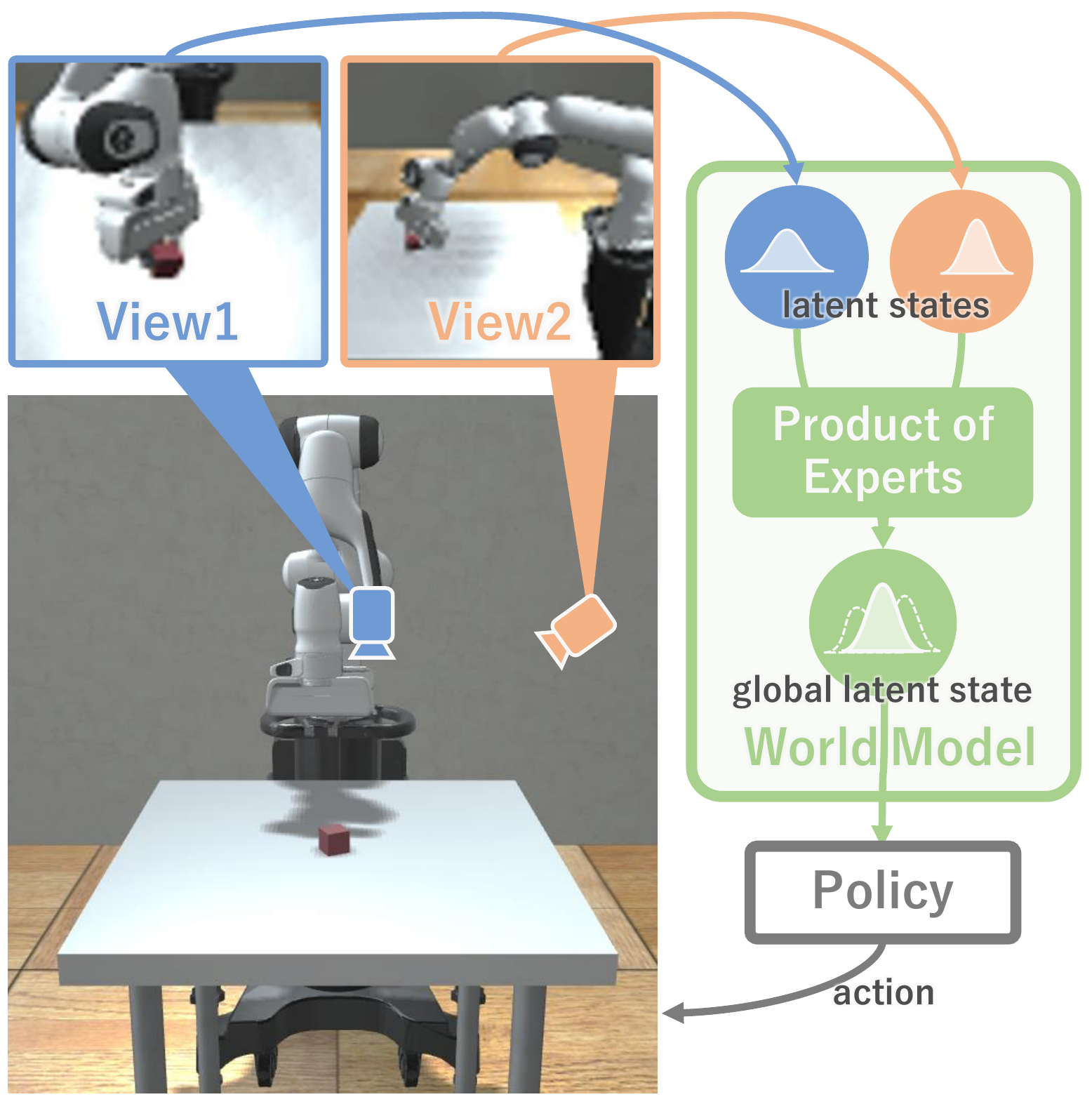}
  \caption{
    Overview of Multi-View Dreaming, the proposed world models approach.
    Multi-View Dreaming trains a shared latent space between different viewpoints using contrastive learning.
    Then, Multi-View Dreaming infers the global latent state by the Product of Experts for multiple latent state distributions.
    By using the global latent state as input observations, the agent can train a policy based on multiple viewpoint observations through reinforcement learning. 
    }
  \label{Overview of Multi-View Dreaming}
  \vspace{-3mm}
\end{figure}

It would be desirable to have a vision-based control system that can manipulate objects in environments where there are many blind spots and image observation is limited. 
In the case of a robot grasping an object on a complicated shelf, the robot must be able to control it by observing images from various cameras. 
    
By contrast, most current reinforcement learning method assumes a single-view observation space, and this imposes limitations on the observed data, such as lack of spatial information and occlusions. 
This makes obtaining ideal observational data from the environment difficult, resulting in problems like missing observational data.
This problem has become a bottleneck for real-world robotics applications. 

Therefore, our goal in this research is to realize a method for learning control based on observations from multiple viewpoints. 
When solving this problem, it will be more useful for robot control in factories where multiple cameras can be installed, as well as automatic driving control where viewing information from multiple directions is required. 
Multi-view reinforcement learning can also be applied to research problems such as robustness to sensor degradation and multimodal data fusion.
To address this problem, it is crucial to develop a model-based reinforcement learning method, which enables integrated recognition and control from multi-view observations.

\begin{figure*}[t!]
  \centering
\includegraphics[width=18cm]{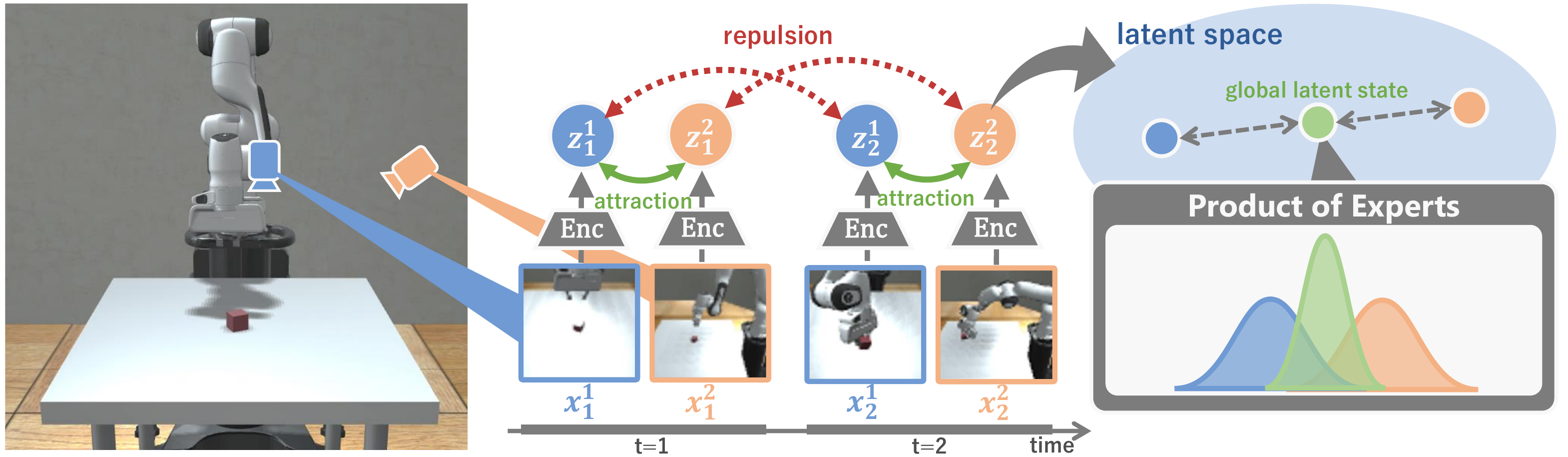}
  \caption{Detailed diagram of multi-view contrastive learning.
  The image pairs of each viewpoint at the same time (green arrows \textcolor[rgb]{0.439,0.678,0.278}{\textbf{$\leftrightarrow$}}) are positive samples, and the latent space is learned so that these images are located close to each other.
  The image pairs of the same and different viewpoints at different times (red arrows \textcolor[rgb]{0.752,0.223,0.211}{\textbf{$\leftrightarrow$}}) are negative samples, and the latent space is learned so that these images are located far from each other.}
  \label{Detailed diagram of Multi-View Contrastive Learning}
  \vspace{-3mm}
\end{figure*}

However, model-based reinforcement learning frameworks for multi-view control have not yet been established.
TCN\cite{sermanet2018time} and mfTCN\cite{dwibedi2018learning}, but both of them do not consider the partial observability of the environment and do not train latent state dynamics.
MuMMI\cite{chen2021multi} is multimodal reinforcement learning, which is highly relevant to this research.
Compared to MuMMI, our focus is to (1) proposing "multi-view" world model which is highly needed in real-world robotics applications, and (2) systematically applying the theory to Dreaming\cite{okada2021dreaming} and DreamingV2\cite{okada2022dreamingv2} to verify its effectiveness.

In this paper, we propose Multi-View Dreaming, a model-based reinforcement learning for control based on multi-view observations. 
Multi-View Dreaming is a novel world model approach for integrated recognition and control from multi-view observations by extending Dreaming.
Fig. \ref{Overview of Multi-View Dreaming} shows an overview diagram of Multi-View Learning. 
We use contrastive learning to train a shared latent space between different viewpoints, and show how the Products of Experts approach can be used to integrate and control the probability distributions of latent states for multiple viewpoints.
\newpage

The proposed method is the extension of Dreaming\cite{okada2021dreaming} and DreamingV2\cite{okada2022dreamingv2} to multi-view control. 
DreamingV2 focuses on representing latent states as categorical variables, while Dreaming focuses on making the dreamer decoder-free. 
The goal of this paper is to develop a multi-view approach to these approaches.

The key contributions of this paper are summarized as follows: 
\begin{itemize}
\item \textbf{Learning world models from multi-view observations.} Using contrastive learning, the proposed Multi-View Dreaming and its categorical version variant, Multi-View DreamingV2, train a shared latent space between different viewpoints. They then use the Product of Experts to infer the global latent state for multiple latent state distributions.
\item \textbf{Practical experiments for visual control.} We demonstrate the effectiveness of the proposed method in some scenarios, which correspond to real-world problems and realistic robot control tasks.
 \end{itemize}

The remainder of this paper is organized as follows.
In Sec.\ref{Related works}, key differences from related work are discussed. 
In Sec. \ref{Multi-View Dreaming}, our proposed method Multi-View Dreaming / Multi-View DreamingV2 is specified. 
In Sec. \ref{Experiments}, the effectiveness of our proposed methods is demonstrated via simulated evaluations. 
Finally, In Sec. \ref{Conclusion}, concludes this paper.


\section{Related works}
\label{Related works}
\subsection*{\bf World Models}
Our research focused on learning world models and policies from high-dimensional observations in partially observable Markov decision process (POMDP \cite{kaelbling1998planning}).
Several approaches have been proposed to learn latent space dynamcis models and use them to solve POMDP in model-based RL \cite{zhang2019solar,kim2019variational}.

World models are a model-based reinforcement learning that uses observation data to learn a predictive model based on the agent's behavior. 
World model trains a latent state dynamics model from the agent's experience, which is used to learn behavior. 
It is advantageous to learn a compact state representation for high-dimensional input information such as images and to use world models to predict the future in latent space.
Representative studies are the World Models\cite{ha2018recurrent}, SLAC\cite{lee2019stochastic}, PlaNet\cite{hafner2018learning}, PlaNet-Bayes\cite{okada2020planet} Dreamer\cite{hafner2019dreamer}, DreamerV2\cite{hafner2020mastering}, Dreaming\cite{okada2021dreaming}, DreamingV2\cite{okada2022dreamingv2},  etc.
Learning from multiple observations is important in real-world robotics applications, but these methods do not address this issue.

The proposed method can also be seen as an extension of the world model to multi-view control, as it can infer the state representation from image observations and predict the future state of the environment in time series using a latent dynamics model.
Our method is especially based on Dreaming\cite{okada2022dreamingv2}.

\subsection*{\bf Contrastive Learning in RL}
Contrastive learning \cite{oord2018representation,chen2020simple} is a self-supervised learning framework for learning useful representations by imposing similarity constraints on the latent space between training data. 

In contrastive learning, the distance of image pairs in the latent space is represented as a loss function.
Contrastive Learning trains image pairs with similarity constraints in the latent space so that they are close to each other if they are data augmented instances, and far from each other if they are different instances.

Works that using contrastive learning for reinforcement learning include CURL\cite{srinivas2020curl}, Dreaming\cite{okada2021dreaming}, CFM\cite{yan2020learning}, CVRL\cite{ma2020contrastive}, TPC\cite{nguyen2021temporal}.

\subsection*{\bf Multi-View Learning in RL}
Several reinforcement learning methods have been proposed to train a policy based on observed data from multiple modalities\cite{lee2019making,li2019multi,chen2021multi,sermanet2018time,dwibedi2018learning}.

MuMMI\cite{chen2021multi} is a research that is particularly relevant to this paper.
In contrast to MuMMI, what is particularly important in this work is that (1) this focuses on "multi-view", which is highly needed in real applications to robots, and (2) systematically applies the theory to Dreamer(V2) and Dreaming(V2) to verify its effectiveness.

TCN\cite{sermanet2018time} and mfTCN\cite{dwibedi2018learning} are examples of research dealing with multi-view contrastive learning, but both of them treating multi-view image embeddings as states, and using them to learn a policy.
However, they do not sufficiently account for the fact that the environment is a POMDP and do not train a latent dynamics model. 
The difference between our method and these studies is that our method is model-based and can predict future states in time series, and it can integrate state representations deduced from multiple viewpoints.


\section{Multi-View Dreaming}
\label{Multi-View Dreaming}
In this paper, we present Multi-View Dreaming, a model-based reinforcement learning method with world models that learns latent dynamics and a policy from multi-view observations, as an extension of Dreaming\cite{okada2021dreaming}.
Fig. \ref{Detailed diagram of Multi-View Contrastive Learning} shows a detailed diagram of the proposed method.
In this method, we apply contrastive learning between multi-view observations to train a world model, based on the idea that images obtained from multi-view observations are augmentations of the same environment instance.
Therefore, images from multi-view observations are trained to be close to each other in latent space at the same time. 
To recognize that observations from different viewpoints have the same latent state, the agent learns world models.

\subsection*{\bf World Model learning based on RSSM}
The world model can learn a predictive model from the agent's experience and use the prediction model to learn the behavior.
Compact state representations are learned when trained on high-dimensional observations as images, allowing forward predictions in the learned latent space.
This kind of model that predicts the future on latent space is called the latent dynamics model. 
By modeling the latent dynamics model of the environment, the agent can predict the long-term future and optimize its behavior without image reconstructions.

Multi-View Dreaming consists of a recurrent state-space model (RSSM) to predict forward dynamics in partially observable environments, and a reward predictor.
RSSM is an important component for learning latent dynamics, and  it has been used in many world models \cite{hafner2019dreamer,hafner2020mastering,okada2021dreaming}.
The model components are:
\begin{equation}
    \begin{split}
        \text { RSSM }\left\{\begin{array}{l} 
        \text {Recurrent model: }\\
        \quad \quad h_{t}=f_\phi \left( h _{t-1}, z _{t-1}, a _{t-1}\right) \\
        \text {Representation model: }\\
        \quad \quad z _{t} \sim q_\phi\left( z _{t} \mid h _{t}, x _{t}\right)\\
        \text {Transition predictor: }\\
        \quad \quad \hat{z}_{t} \sim p_\phi\left(\hat{z}_{t} \mid h _{t}\right) \\
        \end{array}\right. \\
    \end{split}
\end{equation}
\begin{equation}
    \begin{split}
        \text {Reward predictor: }\\
        \quad \quad \hat{r}_{t} \sim p\left(\hat{r}_{t} \mid h _{t}, z _{t}\right) \\
    \end{split}
\end{equation}
\begin{equation}
    \begin{split}
        &\text { Actor: }\quad \hat{a}_{t} \sim p_{\psi}\left(\hat{a}_{t} \mid \hat{z}_{t}\right)\\
        &\text { Critic: } \quad v_{\xi}\left(\hat{z}_{t}\right) \approx \mathrm{E}_{p_{\phi}, p_{\psi}}\left[\sum_{\tau \geq t} \hat{\gamma}^{\tau-t} \hat{r}_{\tau}\right]
    \end{split}
\end{equation}

\begin{figure}[t!]
  \centering
  \includegraphics[width=8cm]{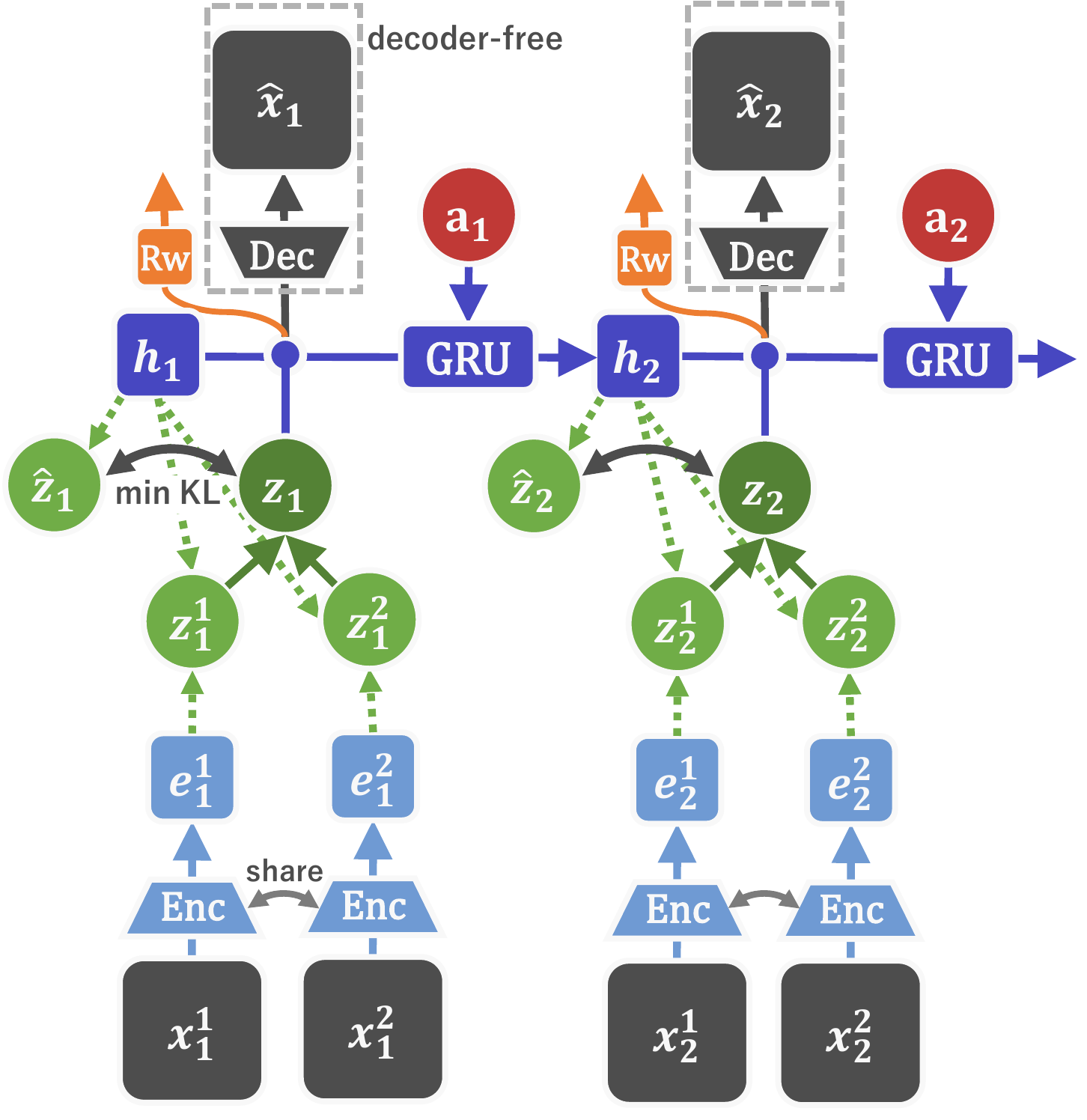}
  \caption{Model Architecture of Multi-View Dreaming.
    In this figure, assume that the model observes images from two viewpoints at the same time.
    The training image $x^{1}_t$, $x^{2}_t$ for each viewpoint is encoded using a shared encoder.
    The RSSM uses a sequence of deterministic recurrent states $h_t$ .
    At each step, this model infers global posterior probability states $z_t$ and prior probability states $\hat{z}_t$.
    The representation model infers posterior probability states $z^{1}_t$ and $z^{2}_t$ for each viewpoint from current images $x^{1}_t$, $x^{2}_t$ for each viewpoint and recurrence states $h_t$.
    The global posterior probability state $z_t$ is calculated from the posterior probability states $z^{1}_t$ and $z^{2}_t$ of each viewpoint by Product of Experts.
    The Transition predictor calculates $\hat{z}_t$, a prior probability state that  attempts to predict the posterior probability state without accessing the current image.
    This method uses the same decoder-free world model as Dreaming, but we train the decoder experimentally without computing the gradient to the loss function.
    }
  \label{Model Architecture of Multi-View Dreaming.}
  \vspace{-3mm}
\end{figure}

Fig. \ref{Model Architecture of Multi-View Dreaming.} illustrates the detailed model architecture of Multi-View Dreaming.
In the proposed model, latent states of the multiple viewpoints $z^{1}_t, z^{2}_t$ are integrated (details in the next section) to global stochastic latent state $z_t$.

\begin{figure*}[t!]
  \centering
  \includegraphics[width=18cm]{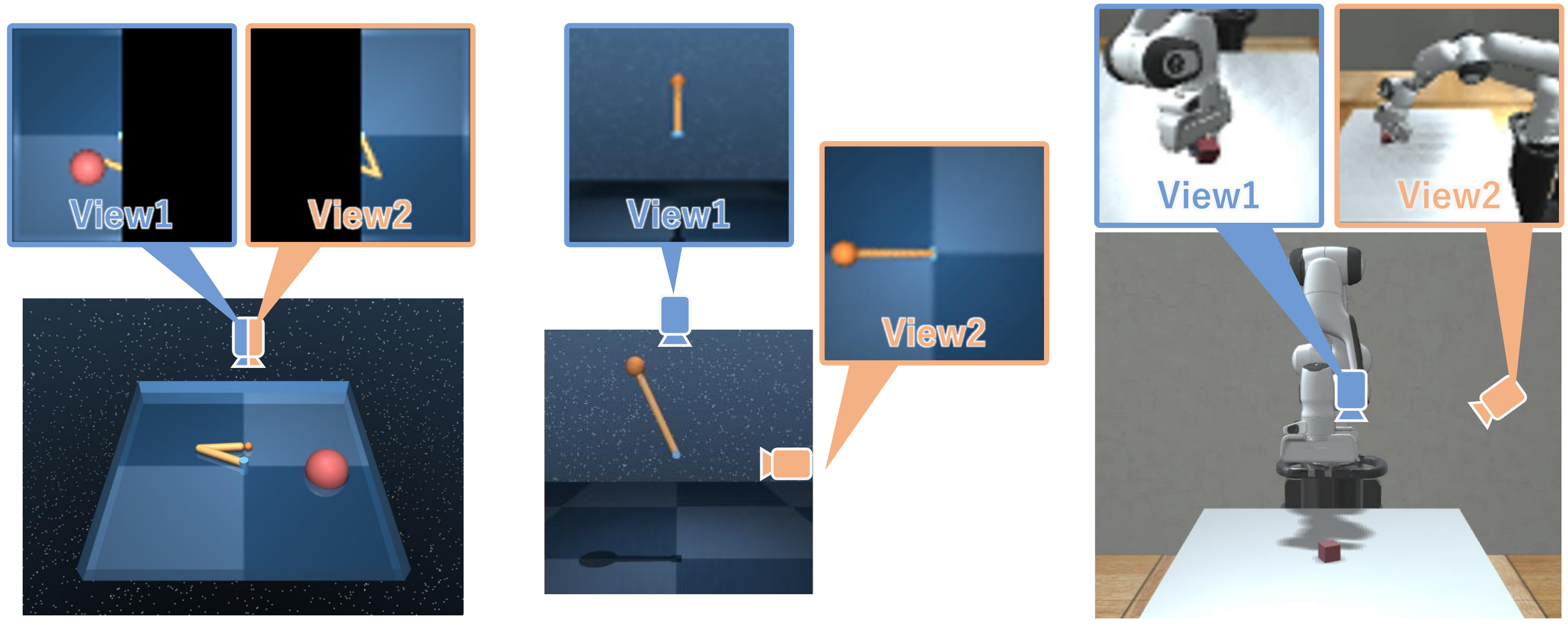}
  \caption{Reacher-easy task uses a continuous action space with 2 dimensions \textbf{(left)}. viewpoint 1 blinds the right half of the screen, and viewpoint 2 blinds the left half. Pendulum-swingup task uses a continuous action space with 1 dimensions \textbf{(center)}. viewpoint 1 observes the pendulum from directly above, and viewpoint 2 observes the pendulum from directly beside. Lift-Panda task uses a continuous action space with 7 dimensions \textbf{(right)}. viewpoint 1 observes the table from the front, and viewpoint 2 observes the table from the side.}
  \label{experiment}
  \vspace{-3mm}
\end{figure*}

\subsection*{\bf Integration of latent state distributions}
In this section, we explain how to integrate multiple latent state distributions.

\subsubsection{\textbf{Multi-View Dreaming (Gaussian)}}
The RSSM based on Dreamer assumes a Gaussian distribution for the latent state distribution.
By integrating the latent states of each of the multiple viewpoints into the global latent state, the global latent state can be seen as representing the true latent state of the environment.
The stochastic state $z_t$ integrates the states of multiple viewpoints by taking a weighted harmonic mean over the mean $\mu$ and variance $\sigma$ of the normal distribution, as shown in the following equation:
\begin{equation}
    \mu_V=\frac{\sum^V_{v=1}{\frac{\mu_v}{\sigma_v^{2}}}}{\sum^V_{v=1}\frac{1}{\sigma_v^{2}}},\;\;
    \sigma_V^2=\frac{1}{\sum^V_{v=1}\frac{1}{\sigma_v^{2}}}
\end{equation}
where $V$ denotes the number of viewpoints.

It is inspired by the Products of Experts\cite{hinton2002training} proposed by Hinton.
The idea is to multiply the density functions of multiple probability distributions (experts) to combine them.
In this way, the agent can choose an action based on several dimensions without covering the full dimensionality of the latent state.

\subsubsection{\textbf{Multi-View DreamingV2 (Categorical)}}
In this paper, we also propose a variant of Multi-View Dreaming that uses a categorical distribution to model the latent state instead of the Gaussian distribution that was proposed in DreamerV2. 
There has been no prior research that uses categorical distributions for latent states of the world model to learn multi-view observations to our knowledge. 
In this paper, we call this method Multi-View DreamingV2. Multi-View DreamingV2 is based on DreamingV2 instead of Dreaming and is extended to multi-view observations. 
Because the latent state in Multi-View DreamingV2 is categorical rather than Gaussian, the Product of Experts is calculated by averaging each dimension of the categorical distribution.

\subsection*{\bf Multi-View Contrastive Learning}
As described in the previous section, to integrate the latent space in multiple viewpoints, it is necessary to learn a common world model across all viewpoint.
We propose to learn a common world model for all viewpoints by using contrastive learning between viewpoints for representation model.

The objective function of Multi-View Dreaming is basically the same as that of Dreaming\cite{okada2021dreaming}.
Dreaming introduces a reconsruction-free objective derived from the ELBO objective:
\begin{equation}
\mathcal{J^{\mathrm{Dreaming}}}:=\sum_{k=0}^{K}\left(\mathcal{J}_{k}^{\mathrm{NCE}}+\mathcal{J}_{k}^{\mathrm{KL}}\right)
\end{equation}
where $K$ represents the overshooting distance.
$J^{KL}_k$ is a multi-step objective and $J^{NCE}_k$ is a categorical cross entropy objective to discriminate positive pair $(z_t,x_t)$ and negative pair $(z_{t}, x^{\prime}(\neq x_{t}))$ as shown below:
\begin{equation}
    \begin{aligned}
        & J _{k}^{ NCE }:= \\
        & E _{\tilde{p}\left( z _{t} \mid z _{t-k}, a _{<t}\right) q\left( z _{t-k} \mid \cdot\right)}\left[\log p\left( z _{t} \mid x _{t}\right)-\log \sum_{ x ^{\prime}} p\left( z _{t} \mid x ^{\prime}\right)\right]
    \end{aligned}
\end{equation}
Dreaming calculates $J^{NCE}_k$ by random image augmentation using image cropping.

In our method, images from each viewpoint observed at the same time are selected in addition to random cropping data augmentation to increase the number of positive sample pairs. 
The negative samples are also sampled from images from different viewpoints observed at different times.
This is based on the intuition that random cropping in contrastive learning can be regarded as equivalent to a change in viewpoint in a reinforcement learning task. 
Even when the viewpoint and observation image are different, we believe that the latent space representations of the same scene should be close together. 
Therefore, by embedding the latent states of different viewpoints at the same time in close proximity, these integrated latent states will be closer to the true latent states. We call this approach as multi-view contrastive learning.

As shown Fig.\ref{Detailed diagram of Multi-View Contrastive Learning}, the image pairs of each viewpoint at the same time are positive samples, and positive samples are image pairs taken from each viewpoint at the same time, and the latent space is learned so that these images are close to each other. 
Negative samples are image pairs of the same and different viewpoints at different times, and the latent space is learned so that these images are far apart.


\section{Experiments}
\label{Experiments}
As shown Fig. 4, we evaluated Multi-View Dreaming's effectiveness in two scenarios that mimic real-world problems. We also used Robosuite to demonstrate the proposed method's effectiveness in a real-world robot control task.

\subsection*{\bf Experimental Settings}

Our main baselines are Dreamer\cite{hafner2019dreamer}, DreamerV2\cite{hafner2020mastering}, Dreaming\cite{okada2021dreaming}, DreamingV2\cite{okada2022dreamingv2}, the representative model-based reinforcement learning methods.
However, we did not simply compare the single-view and multi-view approaches, but also extended the baseline as follows: a simple extension of the baseline by overlaying multi-view images in the color channel direction as input images.
For example, if the image of the 64(height)$\times$64(width)$\times$3(color) array is observed from two viewpoints, the agent will observe the image of the 64(height)$\times$64(width)$\times$6(color) array after overlaying.
Multi-View Dreaming is implemented based on DreamingV2.
Therefore, the elements proposed in the research up to DreamingV2 will be inherited in this method.

We experimented with the three tasks shown in Fig. 4.
In all tasks, observations are pixel inputs (64$\times$64) only.
Reacher task and Pendulum task are provided from the DeepMind Control Suite\cite{deepmindcontrolsuite2018}, Lift task is provided from the Robosuite\cite{zhu2020robosuite}, but environments were augmented to provide images from multiple viewpoints.

\begin{figure}[t!]
  \centering
  \includegraphics[width=7cm]{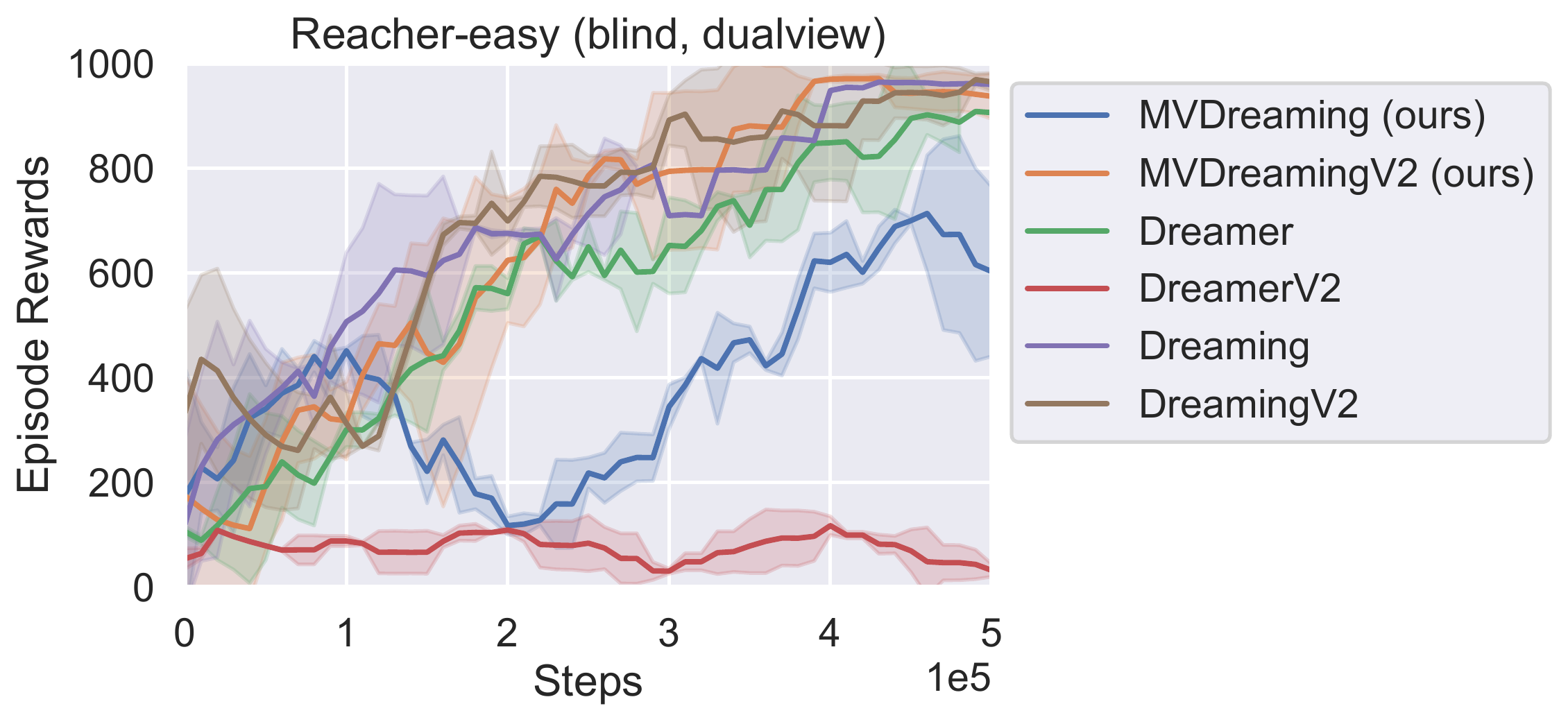}
  \caption{Training progress of Blind Reacher scenario}
  \label{Reacher}
  \vspace{-3mm}
\end{figure}
\begin{figure}[t!]
  \centering
  \includegraphics[width=7cm]{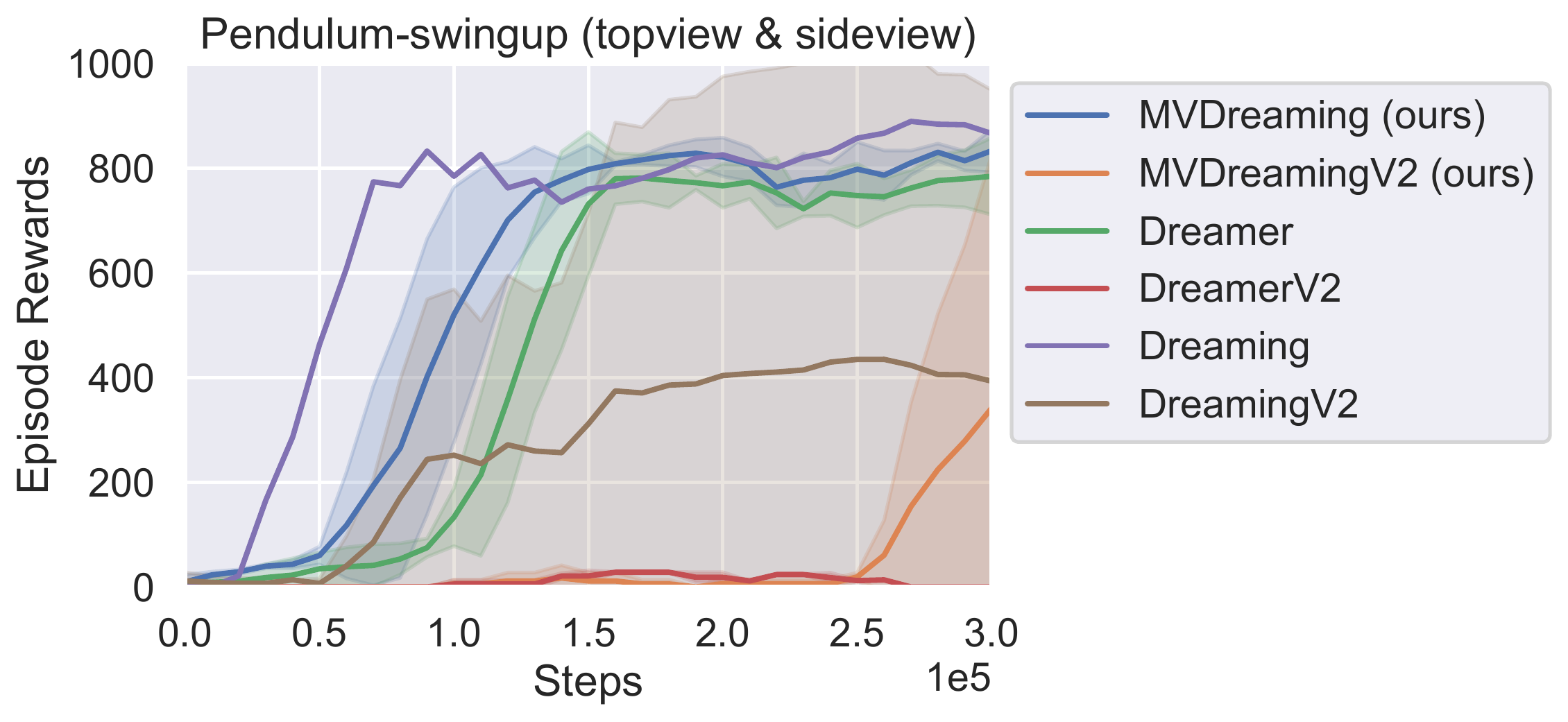}
  \caption{Training progress of Dual View Pendulum scenario}
  \label{Pendulum}
  \vspace{-3mm}
\end{figure}
\begin{figure}[t!]
  \centering
  \includegraphics[width=7cm]{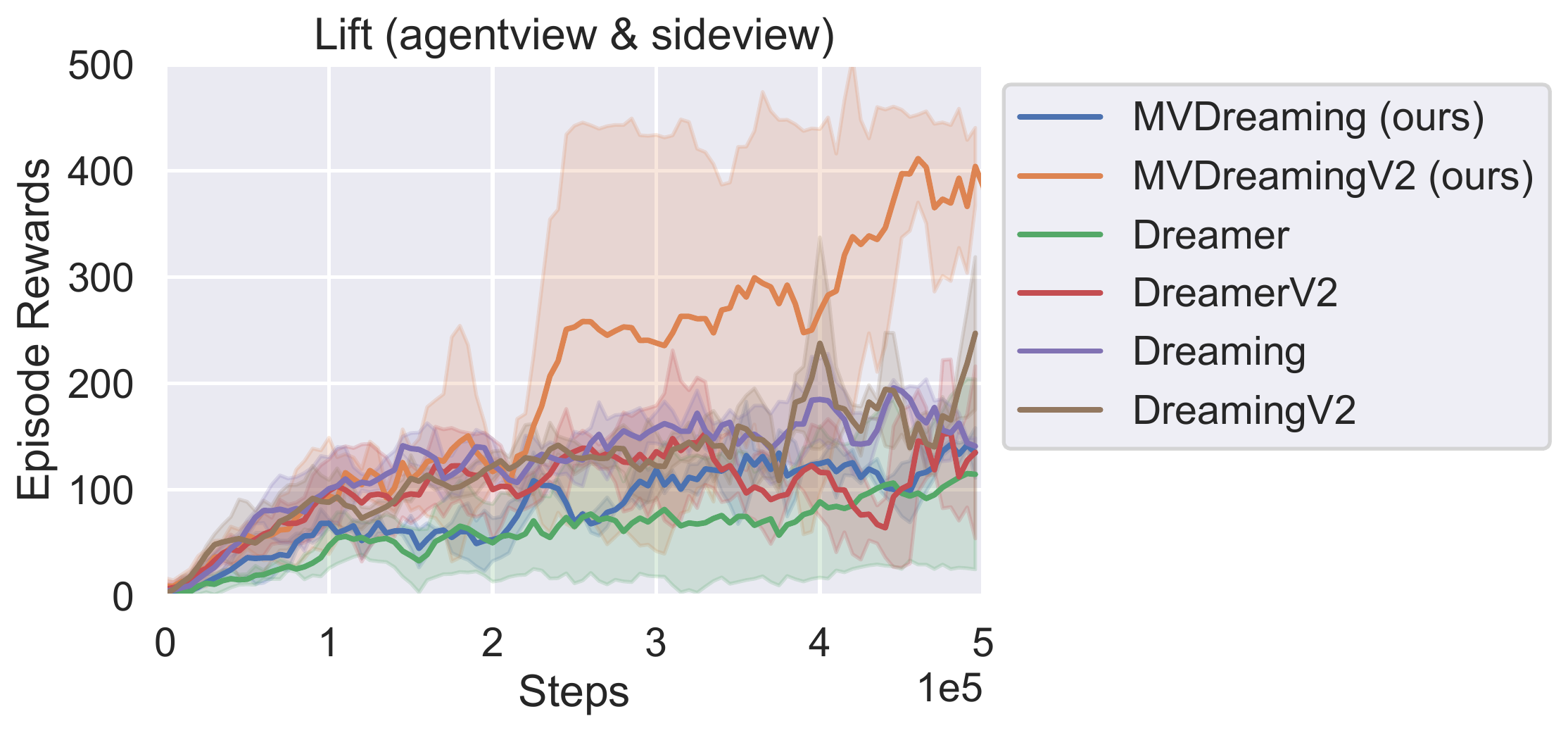}
  \caption{Training progress of Robosuite Lift scenario}
  \label{Lift}
  \vspace{-3mm}
\end{figure}
\begin{figure}[t!]
  \centering
  \includegraphics[width=7cm]{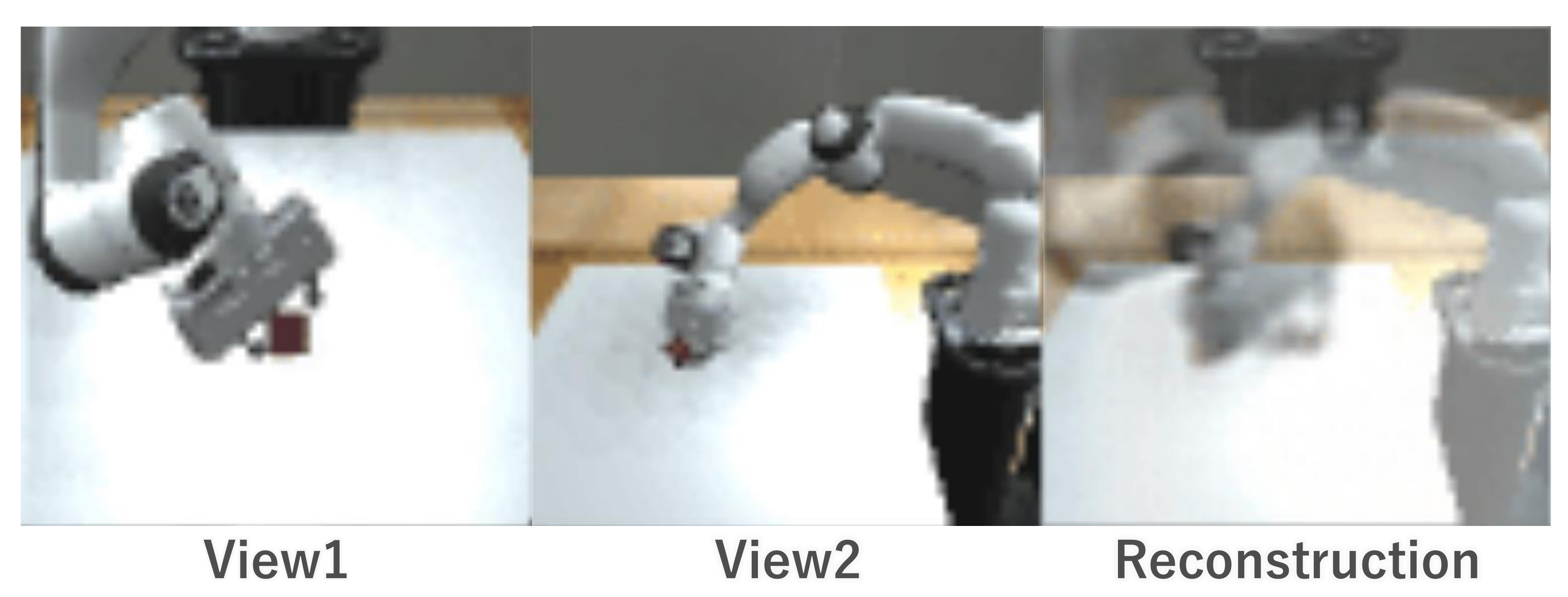}
  \caption{Observed image from viewpoint 1 \textbf{(left)}. Observed image from viewpoint 2 \textbf{(center)}. Reconstructed image from the global latent state \textbf{(right)}. }
  \label{reconstruction}
  \vspace{-3mm}
\end{figure}

\begin{table*}[ht]
\vspace{4mm}
\caption{Comparison of Multi-View Dreaming / V2 to simple extension of conventional model-based reinforcement learning by final performance score.}
\begin{center}
\begin{small}
\begin{tabular}{lcccccccc}
\toprule
 & MVDreaming & MVDreamingV2 & Dreamer & DreamerV2 & Dreaming & DreamingV2 \\
 & (ours) & (ours) & \cite{hafner2019dreamer} & \cite{hafner2020mastering} & \cite{okada2021dreaming} & \cite{okada2022dreamingv2} \\
\midrule
Multi-View & $\checkmark$ & $\checkmark$ & & & & \\
Gaussian Latent & $\checkmark$ & & $\checkmark$ & & $\checkmark$ \\
Categorical Latent & & $\checkmark$ & & $\checkmark$ & & $\checkmark$ \\
Decoder-free & $\checkmark$ & $\checkmark$ & & & $\checkmark$ & $\checkmark$ \\
\midrule
Reacher (multi-view) &588.6$\pm$356.0 &\textbf{936.1$\pm$95.9} & 685.0$\pm$410.7 &42.2$\pm$73.3  &841.3$\pm$316.7  & 860.9$\pm$285.6 & \\
Pendulum (multi-view) & 812.2$\pm$130.4 & 256.5$\pm$304.5 & 801.2$\pm$110.4 & 10.8$\pm$24.5 &\textbf{831.6$\pm$126.4} & 410.4$\pm$413.99 &  \\
Lift (single-view)  & --- & --- & 133.5$\pm$64.28 & 132.6$\pm$62.4 & 150.5$\pm$78.5& \textbf{330.9$\pm$113.6} &  \\
Lift (multi-view)   &110.1$\pm$44.6 & \textbf{345.0$\pm$133.6} & 102.9$\pm$82.8 & 120.8$\pm$51.58& 177.0$\pm$80.9& 254.7$\pm$104.2 &  \\
\midrule
\bottomrule
\end{tabular}

\end{small}
\end{center}
\vskip -0.1in
\end{table*}

\subsection*{\bf Scenario: Blind Reacher}
In this experiment, we assume that occlusion occurs on the observed images and that the critical information required for the task is not available from a single viewpoint. 
Specifically, as Fig. 4 shows, viewpoint 1 blinds the right half of the screen, and viewpoint 2 blinds the left half. We handle blinds by filling them with black pixels, so the image size is unchanged. 
In this scenario, learning the policy from a single viewpoint image is insufficient to complete the task; instead, the policy must be learned by combining information from multiple viewpoints.

\subsection*{\bf Scenario: Dual View Pendulum}
In this experiment, we assume a scenario where the camera position in the environment is changed from the default and only limited information of the task is available from a single viewpoint. Viewpoint 1 observes the pendulum from directly above, while viewpoint 2 observes the pendulum from directly beside, as shown in Fig. 4. It is difficult to achieve any of these viewpoints with only a single camera in one direction, and it is necessary to learn the policy by combining information from two viewpoints located in different places.

\subsection*{\bf Scenaro: Robosuite Lift}
We used Robosuite in this experiment to test the effectiveness of our method in a realistic robot control task. In a realistic robot task, the robot's body and arms act as obstacles in the observation space, necessitating multi-modal control from multiple viewpoints.

\subsection*{\bf Experimental Results}

Fig. \ref{Reacher}, \ref{Pendulum}, \ref{Lift} show the training progress for each scenario, and the final performance scores are shown in Table 1. 

\textbf{In the Blind Reacher scenario}, Multi-View DreamingV2 learned policies by using the images from the two viewpoints well.
However, the performance was comparable to simple extensions of Dreamer, Dreaming, and DreamingV2. 
Multi-View Dreaming did not perform as well as the full observation.

\textbf{In the Dual View Pendulum scenario}, all methods except DreamerV2 and Multi-View DreamingV2 performed equally well in learning. 

\textbf{In the Robosuite Lift scenario}, Multi-View DreamingV2 outperformed all other approaches by a significant difference.
Whereas the previous two experiments did not differ from baseline, Multi-View DreamingV2 was particularly effective in this task.

This suggests that it is difficult to separate the necessary information for a task like Lift, which involves complex image information and different dynamics between viewpoints, using a simple method of overlaying images in the color channel direction, and that the proposed method is effective in estimating the latent state for each viewpoint.
A reconstructed image from the global latent state is shown in Fig. 8. The global latent state combines and embeds both the important information of viewpoint 1 and viewpoint 2. This can be qualitatively confirmed.

\section{Conclusion}
\label{Conclusion}
In this paper, we proposed a novel world model approach for integrated recognition and control from multi-view observations by extending Dreaming.
We used contrastive learning to train a shared latent space between different viewpoints, and showed how the Products of Experts approach can be used to integrate and control the probability distributions of latent states for multiple viewpoints.

We demonstrated the effectiveness of the world model using multi-view observations in two scenarios that correspond to problems in real environments and in realistic robot control tasks.
In conclusion, simple extensions of methods of overlapping images are effective for simple tasks, but a multi-view contrastive learning approach is more effective for tasks with complex images and dynamics. 
Understanding the features of these methods revealed in this study, as well as making effective use of multi-view, is important for practical applications in robotics. 
Theoretical causes of these differences will be the subject of future research. 

Our method can be seen as an example of a multi-view version of the generalized multimodal world model.
In addition to multi-view images, world models that incorporate more modalities such as audio, tactile sensing, and depth sensors would be an intriguing one which could be usefully explored in further research.
In addition, embedding domain information such as camera coordinates and robot proprioception into the latent state of each viewpoint would be a fruitful area for further work.

We were not able to experiment with the real robot in this paper, but we are currently working on real robotics application, and evaluating it will be a future issue.

\bibliography{iros2022}
\bibliographystyle{ieeetr}
\end{document}